\pdfoutput=1

\documentclass[11pt]{article}

\usepackage[]{ACL2023}

\usepackage{times}
\usepackage{latexsym}
\usepackage{graphicx}
\usepackage{adjustbox}
\usepackage[noend]{algpseudocode}
\usepackage{algorithmicx,algorithm}
\usepackage[T1]{fontenc}

\usepackage{amsmath,graphicx}

\usepackage[utf8]{inputenc}

\usepackage{microtype}

\usepackage{inconsolata}

\title{AMTSS: An Adaptive Multi-Teacher Single-Student Knowledge Distillation Framework For Multilingual Language Inference}

\author{
Qianglong Chen\textsuperscript{1,2}
\thanks{\quad Work is done during internship at Alibaba Group.}, Feng Ji\textsuperscript{3}\thanks{\quad The work is mainly conducted while being at Alibaba Group.}, Feng-Lin Li\textsuperscript{2}, Guohai Xu \textsuperscript{2}, Ming Yan \textsuperscript{2}\\ \textbf{Ji Zhang\textsuperscript{2}, Yin Zhang\textsuperscript{1}\thanks{\quad Corresponding Author: Yin Zhang}} \\
  \textsuperscript{1}College of Computer Science and Technology, Zhejiang University, China \\
  \textsuperscript{2}DAMO Academy, Alibaba Group, China, \textsuperscript{3}Tencent, China \\
  {\tt 
  \{chenqianglong,zhangyin98\}@zju.edu.cn}, {\tt maillifenglin@gmail.com}\\ 
  {\tt 
  \{guohai.xgh,ym119608,zj122146\}@alibaba-inc.com}, 
  {\tt neilji@tencent.com}\\
}

\begin{document}
\maketitle
\begin{abstract}
Knowledge distillation is of key importance to launching multilingual pre-trained language models for real applications. To support cost-effective language inference in multilingual settings, we propose \textbf{AMTSS}, an \textbf{a}daptive \textbf{m}ulti-\textbf{t}eacher  \textbf{s}ingle-\textbf{s}tudent distillation framework, which allows distilling knowledge from multiple teachers to a single student. We first introduce an adaptive learning strategy and teacher importance weight, which enables a student to effectively learn from max-margin teachers and easily adapt to new languages. 
Moreover, we present a shared student encoder with different projection layers in support of multiple languages, which contributes to largely reducing development and machine cost. Experimental results show that AMTSS gains competitive results on the public XNLI dataset and the realistic industrial dataset AliExpress (AE) in the E-commerce scenario.
\end{abstract}

\section{Introduction}
\label{sec:intro}
Multilingual pre-trained language models (aka M-PLMs) such as mBERT ~\cite{devlin-etal-2019-bert}, XLM ~\cite{conneau2019unsupervised} and XLM-R ~\cite{conneau-etal-2020-unsupervised} have achieved significant improvement for many multilingual tasks, including multilingual NLI~\cite{conneau-etal-2018-xnli,bowman-etal-2015-large,williams-etal-2018-broad}, question answering~\cite{lewis2019mlqa,clark-etal-2020-tydi,hardalov2020exams} and NER~\cite{wu-dredze-2019-beto,rahimi-etal-2019-massively}.
As pre-trained language models are usually computationally expensive, many transformer distillation methods~\cite{jiao-etal-2020-tinybert,liu2020cross} have been proposed, which distill knowledge from a large teacher model to a lightweight student network, to accelerate inference and reduce model size while maintaining the accuracy.

The majority of works mainly focus on learning from a single teacher as in Figure~\ref{fig:pipeline} (a), while only a few studies have considered to learn from multiple teachers~\cite{peng2020mtss,liu2020adaptive} which allows to select the optimal model as teacher for different domains during the student training. 
It is essential for cross domain knowledge distillation, especially in multilingual NLI.


In this work, we focus on chatbot settings, which currently supports nearly twenty kinds of languages on E-commerce language inference, and are constantly accepting new languages.
Considering the linearly increased development and machine cost, we cannot develop a model instance for each language, which has poor scalability.
Meanwhile, as we need to distill from around twenty teacher models each time when dealing with a new coming language, current multi-teacher distillation methods~\cite{peng2020mtss,liu2020adaptive,10.1145/3097983.3098135,yang2020model} are not fit for our scenario. 
This begs an important question in practice: \textit{can we distill knowledge from multiple teachers in a multilingual settings to cost-effectively support multiple languages and easily adapt to a new language?}

To address this challenge, we propose an adaptive multi-teacher single-student distillation framework (AMTSS). 
Firstly, we fine-tune a pre-trained language model for each language and obtain the optimal teacher, either monolingual or multilingual. 
Then, we distill the knowledge from teachers to a single student with a novel adaptive training strategy and a shared student encoder with different projection layers, instead of training several students for each language. 
For adapting to the new languages, we fine-tune the student model to learn from the max-margin teachers instead of re-training the student model with all teachers.

The contributions of this work are as follows:
\begin{itemize}
    \item We propose an adaptive multi-teacher single-student knowledge distillation framework, consists of a shared student encoder with different projection layers in support of multiple languages in a cost-effective manner.
    \item We propose a weight based adaptive learning strategy that enables a student model to effectively learn from max-margin teachers with the importance weights, and easily adapt to new coming languages.
    \item We demonstrated the effectiveness of AMTSS through the experimental evaluation on public XNLI dataset and a realistic industrial dataset AliExpress (AE) in E-commerce scenario.
\end{itemize}

\section{Knowledge Distillation Framework}
\begin{figure}[htp]
    \centering
    \includegraphics[width=1\linewidth]{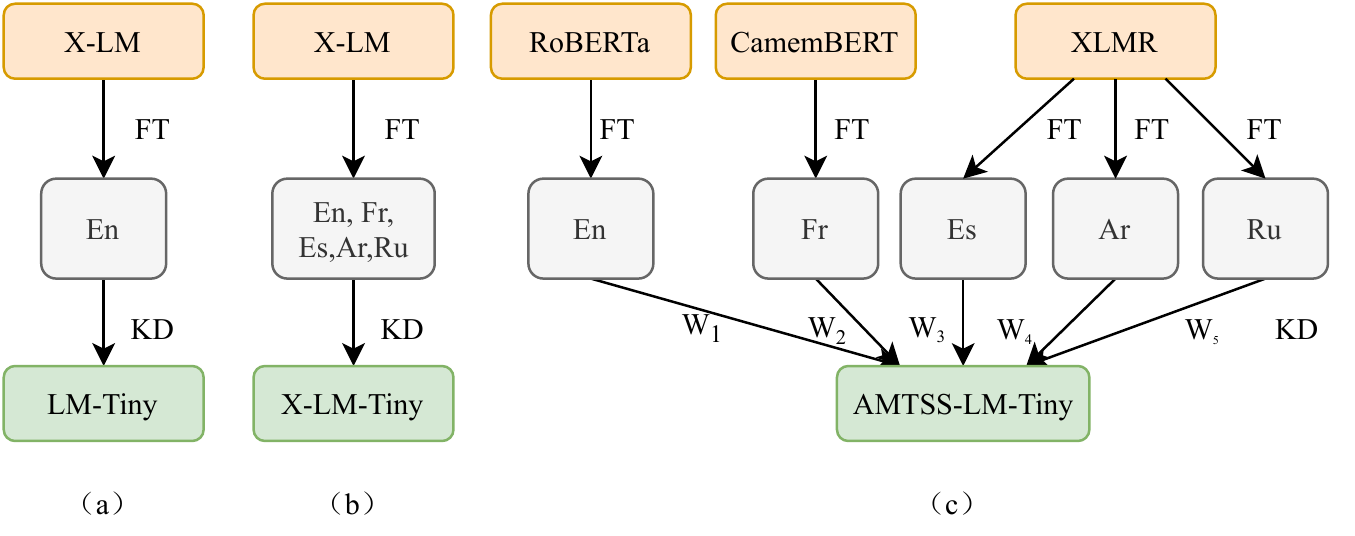}
    \caption{Knowledge distillation framework: a) monolingual LM knowledge distillation, b) original multilingual LM knowledge distillation,
    c) adaptive-MTSS-LM distillation. FT denotes fine-tuning, KD denotes knowledge distillation, $W_i$ is weighted parameter.}
    \label{fig:pipeline}
\end{figure}

As shown in Figure~\ref{fig:pipeline} (c), we present an adaptive knowledge distillation framework. 
Being different from standard knowledge distillation, either monolingual in Figure~\ref{fig:pipeline} (a) or multilingual in Figure~\ref{fig:pipeline} (b), AMTSS enables a student to not only learn from multiple teachers with different importance weights, but also cost-effectively serve multiple languages and easily adapt to new coming language.

\subsection{Model Architectures}
\subsubsection{Teacher models} 
To ensure teacher models achieve the best performance on each language, we adopt different architectures (e.g., CamemBERT or XLMR).
That is, one can choose different pre-trained LM, monolingual or multilingual, together with a simple projection layer, to fine-tune and obtain an optimal teachers for a specific language.
We formulate teacher models as follows:
\begin{equation}
    input = [D_1,...,D_i,...,D_n] 
\end{equation}
\begin{equation}
    h^T_i = encoder^T_i(input_i)
\end{equation}
\begin{equation}
    \hat{y}^T_{t,i} = softmax^T_{t,i}(h^T_i)
\end{equation}
\begin{equation}
    \hat{y}^T_i = \sum_{t=1}^{n} w_{t,i} \hat{y}^T_{t,i}
\end{equation}
where $input_i$ denotes the input of the $i$-th language, $h^T_i$ is the output hidden state of $i$-th teacher,$T$ denotes teacher, $softmax^T_i$ is the specific prediction layer of $i$-th teacher model, $w_{t,i}$ denotes the importance weight of $i$-th teacher and $\hat{y}^T_i$ represents the soft-targets generated by the $i$-th teacher.

\subsubsection{Student model} 
For student model, we use a shared 4-layer transformer with different projection layers for each language. Specifically, each $input_i$ is encoded by the shared student encoder (Equation~\ref{equ:student_encoder}), and the prediction is generated by a corresponding projection layer for each language (Equation~\ref{equ:student_logits}): 
\begin{equation}
    h^S_i = encoder^S(input_i)
    \label{equ:student_encoder}
\end{equation}
\begin{equation}
    \hat{y}^S_i = softmax^S_i(h^S_i)
    \label{equ:student_logits}
\end{equation}
where $S$ denotes student, $h^S_i$ is the output hidden state of student encoder, $\hat{y}^S_i$ represents the soft-targets generated by the student.
Note that the designing of shared encoder plus different projections plays a key role in supporting multiple languages with one lightweight student model, which contributes to largely reducing development and machine cost.

\subsection{Training Loss}
For loss function, we use cross-entropy to measure the difference between the output distribution of student and ground truth, and use KL-divergence loss to measure the distance between the output of student and that of teachers, defined as follows:
\begin{equation}
    \label{loss}
    L_{KD} = \sum_{i} H(y_i,y_i^S) + \lambda \sum_{i} D_{KL}(\hat{y}_i^T,\hat{y}_i^S)
\end{equation}
where $H$ is the cross-entropy loss, $y_i$ and $y_i^S$ are the ground-truth and inferred labels of student, $D_{KL}$ is the KL divergence, $\hat{y}_i^T$ and $\hat{y}_i^S$ denote the soft-targets generated by teacher and student model, $\lambda \in (0,1)$ is a hyper-parameter that controls the relative influence of teacher knowledge transfer.

\begin{algorithm}[!ht]
\caption{\textbf{Adaptive Training Strategy}}
\label{algo:algorithm}
\hspace*{0.02in} {\bf Input:}
Teachers $T_1, T_2, ..., T_n$, Weights $W_1,W_2,..., W_n$, Datasets $\{D_1,...,D_i,...,D_n\}$\\
\hspace*{0.02in} {\bf Output:}
Student Model $S$
\begin{algorithmic}[1]
\State $D_{i} \gets \{D_1,...,D_n\}$
\State $D_{i}$ consists of $\{D_{train,i},D_{valid,i}\}$
\For{$ epoch < M $}
\For{$D_{train,i}\in D_i$}
\State Training $S$ with weighted teachers, and $D_{train,i}$
\State Update parameters  $W_1,W_2,..., W_n$
\EndFor
\For{$D_{valid,i}\in D_i$}
\State Evaluate($D_{valid,i}$,$S$)
\State $M_i \gets Margin(S,W_i T_i)$
\EndFor
\State $R \gets Rank(M_1,...,M_n)$
\State $D_{i} \gets TopK(index(R))$
\If{$max(R) <\epsilon$}
\State break
\EndIf
\EndFor
\end{algorithmic}
\end{algorithm}

\subsection{Adaptive Training Strategy}

To make student models easily adapt to new languages, we introduce an adaptive training strategy with max-margin and importance weight, as in Algorithm~\ref{algo:algorithm}. Given $n$ languages, we firstly obtain $n$ teachers $T_1, T_2, ..., T_n$ through fine-tuning on the corresponding datasets $D_1, D_2, ...,D_n$. For the knowledge distillation, we obtain a student model $S$ through learning from the $n$ importance weighted teachers (line 4-6). Further, we evaluate $S$ on the $n$ validation datasets (line 7-9) and calculate the top-K max-margin~\footnote{The margin denotes the performance delta between student and a teacher.} teachers (line 10). In general, the distillation process will continue to learn from the chosen top-K max-margin teachers (line 11), and will terminate if the number of epoch exceeds $M$ or the max-margin is smaller than a certain threshold $\epsilon$ (line 12-13). Our adaptive strategy is inspired by Adaboost~\cite{freund1997decision} that tries to improve model performance through learning from mistaken instances, but differs from that we only train one base model instead of a group, and do not update the importance weight of teachers. Given a new language, we can directly continue our distillation process by treating the newly obtained teacher as a max-margin teacher. Through the importance weight learning, we aim to obtain a balanced student model from multiple teachers.

\section{Experiments and Results}

\subsection{Datasets} We evaluated our framework on two datasets: \textbf{XNLI} and \textbf{AE}. XNLI~\cite{conneau-etal-2018-xnli} is a public multilingual NLI dataset, which the number of categories in XNLI is 3, including entailment, contradiction and neutral.
The AliExpress (AE) dataset is a practical text classification dataset constructed from our chatbot AliExpress in E-commerce scenario. 
We select 5 languages from AliExpress chatbot for evaluation, and the number of categories for each language is 20.
The statistics of the practical AE dataset label distribution are shown in Appendix.
Since the label distribution are not equal for difference language set, it is challenging for current knowledge distillation methods.
Therefore, we propose the importance weight based adaptive strategy for knowledge distillation.

\begin{table*}[!ht]
    \centering
    \caption{Results on XNLI, including teacher models, X-LM-Tiny and Adaptive MTSS-LM-Tiny. The evaluation metric is accuracy. On XNLI, the performance of XLM-R on En and Fr is 89.1\% and 83.5\%, lower than RoBERTa and CamemBERT.}
    \begin{tabular}{|l|c|c|c|c|c|c|}
    \hline
      \textbf{Model} & \textbf{Ar} & \textbf{En} & \textbf{Es} & \textbf{Fr} & \textbf{Ru} & \textbf{Average} \\
    \hline
     Teacher Architecture & XLM-R & RoBERTa  & XLM-R & CamemBERT & XLM-R & - \\
    \hline
       Teacher & 83.10 & 91.30 & 86.60 & 85.10 & 83.50 & 85.92 \\
    \hline
       X-LM-Tiny & 79.40 & 87.60 & 80.06 & \textbf{82.50} & 79.90 & 81.88\\
    \hline
        Adaptive MTSS-LM-Tiny   & \textbf{81.42} & \textbf{90.91} & \textbf{82.49} & 82.46 & \textbf{80.82} & \textbf{83.62} \\
    \hline
    \end{tabular}
    \label{tab:XNLI}
\end{table*}

\begin{table*}[!ht]
    \centering
    \caption{Results on the AE dataset, including teacher models, X-LM-Tiny and Adaptive MTSS-LM-Tiny. The evaluation metric is accuracy. On AE, the performance of XLM-R on En and Fr is 83.04\%, and 76.55\%, respectively. }
    \begin{tabular}{|l|c|c|c|c|c|c|}
    \hline
      \textbf{Model} & \textbf{Ar} & \textbf{En} & \textbf{Es} & \textbf{Fr} & \textbf{Ru} & \textbf{Average} \\
    \hline
     Teacher Architecture & XLM-R & RoBERTa  & XLM-R & CamemBERT & XLM-R & -\\
    \hline
       Teacher & 80.45 & 88.92 & 81.05 & 84.17 & 82.55 & 83.43  \\
    \hline
       X-LM-Tiny  & {78.21} & 77.60 & {79.52} & {81.49} & {79.64} & 79.29 \\
    \hline
        Adaptive MTSS-LM-Tiny & \textbf{79.45} & \textbf{89.93} & \textbf{80.69} & \textbf{83.56} & \textbf{81.82} & \textbf{83.09} \\
    \hline
    \end{tabular}
    
    \label{tab:AE NLI}
\end{table*}

\subsection{Experimental Setting}
We use the RoBERTa-large for English Corpus encoding, CamemBERT for French Corpus encoding and XLM-R-large for Spanish, Arabic and Russian encoding. We use AdamW as optimizer and adopt cross entropy and KL divergence as the loss function. We set batch size to 32, the learning rate to 1e-5, dropout to 0.01. For training, we set the max epoch to 200. The evaluation metric is accuracy.


\subsection{Baseline} {X-LM-Tiny} is the student model which is distilled from XLM-R-large through original multilingual LM knowledge distillation as shown in Fig.\ref{fig:pipeline}(b), where XLM-R-large is a single teacher trained with the whole mixed multilingual dataset. We use X-LM-Tiny as the baseline to evaluate the effectiveness of our adaptive MTSS framework. 

\begin{table}[!ht]
    \centering
    \caption{The size of teacher and student parameters.}
    \begin{tabular}{|l|c|}
    \hline
        \textbf{Model Name} & \textbf{Size}  \\
    \hline
        XLM-R & 550M \\ 
        RoBERTa & 355M \\
        CamemBERT & 335M \\
        X-LM-Tiny & 52.2M \\ 
        AMTSS-LM-Tiny & 52.2M \\
        
    \hline
    \end{tabular}
    
    \label{tab:size}
\end{table}

\subsection{Results and Analysis}
The results on XNLI are shown in Table~\ref{tab:XNLI}. While the X-LM-Tiny obtain 81.88\% on average, while with the adaptive strategy, our Adaptive MTSS-LM-Tiny can gain 1.74\% improvements on average. The results demonstrate the effectiveness of multiple teachers and adaptive training for knowledge distillation. 
Note that for the teacher architecture on English and French, we adopt RoBERTa and CamemBERT respectively, due to the performance of them are better than XLM-R.

The practical experimental results on AE are shown in Table~\ref{tab:AE NLI}. The performance of our Adaptive MTSS-LM-Tiny is only 0.34\% lower than that of teachers on average, and 2.80\% higher than X-LM-Tiny. 
That is, a student model can benefit from other teachers and outperform its dedicated one. Specifically, compared with X-LM-Tiny, the performance of AMTSS-LM-Tiny is 12.33\% higher on English, and 2.29\% higher on average. The notable gap between X-LM-Tiny and XLM-R on English ($12.33\%=89.93\%-77.60\%$) indicates that the direct mix of multilingual training data could cause unexpected deviations in some languages. With the importance weight based adaptive training strategy, our adaptive MTSS architecture can achieve more balanced improvement on each language, while the performance of X-LM-Tiny in each language has large gap to the teacher. The gaps between X-LM-Tiny with RoBERTa-large in English and with CamemBERT in French are 11.32\% and 2.68\%, while our Adaptive MTSS-LM-Tiny reduce the gap to 1.01\% and 0.61\% respectively. 

We also analyse the model size of both teacher and student models, and report the statistics in Table~\ref{tab:size}. The size of student models is much smaller than that of teachers, which can be deployed on restricted resources (e.g., CPU) at a low cost. Although X-LM-Tiny has the same size of parameters, the performance are not comparable with our Adaptive MTSS-LM-Tiny. Meanwhile, the X-LM-Tiny are not suitable for adapting to new language.

\subsection{Adapting to New Language}
To test the adaptability of our framework to new languages, we introduce another two languages, Korean (Ko) and Polish (Pl), in addition to the aforementioned 5 languages. We train two new teachers and regard them as max-margin teachers to continue the previous distillation process, learning the teachers importance weight, and report the results in Table~\ref{tab:adapt language}. We can find that the performance of student on Ko and Pl is 77.91\% and 81.64\%, which is higher than that of the teachers, and the average performance of all 7 languages decreases less ($83.09\% \rightarrow 82.15\%$).
That is, with adaptive training strategy, we can alleviate the information forgetting, and even benefit from previous language while adapting to new languages.

\begin{table}[!ht]
    \centering
    \caption{The results of AMTSS-LM-Tiny after adapting. }
    \begin{adjustbox}{width=0.48 \textwidth}
    \begin{tabular}{|c|c|c|c|}
    \hline
      \textbf{Model}  &  \textbf{Ko} & \textbf{Pl} & \textbf{All Average}\\
    \hline
       Architecture & XLM-R & XLM-R & -\\
    \hline
        Teacher  & 77.25 & 81.40 & 82.26 \\
    \hline
        AMTSS-LM-Tiny  & 77.91 & 81.64 & 82.15\\
    \hline
    \end{tabular}
    \end{adjustbox}
    
    \label{tab:adapt language}
\end{table}

\section{Conclusion}
In this paper, we propose AMTSS, an adaptive multi-teacher single-student knowledge distillation framework, which enables a student model to learn from multiple teachers with adaptive strategy and importance weight. Experimental results demonstrate that our model is cost-effectively serve multiple languages, and easily adapt to new languages.
In the future, we will further explore adapting max-margin teacher weights with contrastive learning to improve the performance of model and alleviate the data imbalance problem in practical scenarios.

\section*{Limitations}
In this work, although we evaluate our AMTSS methods on both public XNLI dataset and the realistic industrial dataset AliExpress (AE) in E-commerce scenario, the AMTSS method can be further evaluated on other scenarios and tasks, such as question answering, commonsense reasoning in medical or science area. Furthermore, in this work, we explore the possibility of adaptive strategy and importance weight for knowledge distillation, but there are more methods we are going to introduce into knowledge distillation, such as contrastive learning, few-shot learning and in-context learning etc., to alleviate the problem in different low-source language.
\bibliography{anthology,custom}
\bibliographystyle{acl_natbib}


\clearpage
\appendix
\section{Statistical results of AE dataset}

\begin{figure}
    \centering
    \includegraphics[width=1\linewidth]{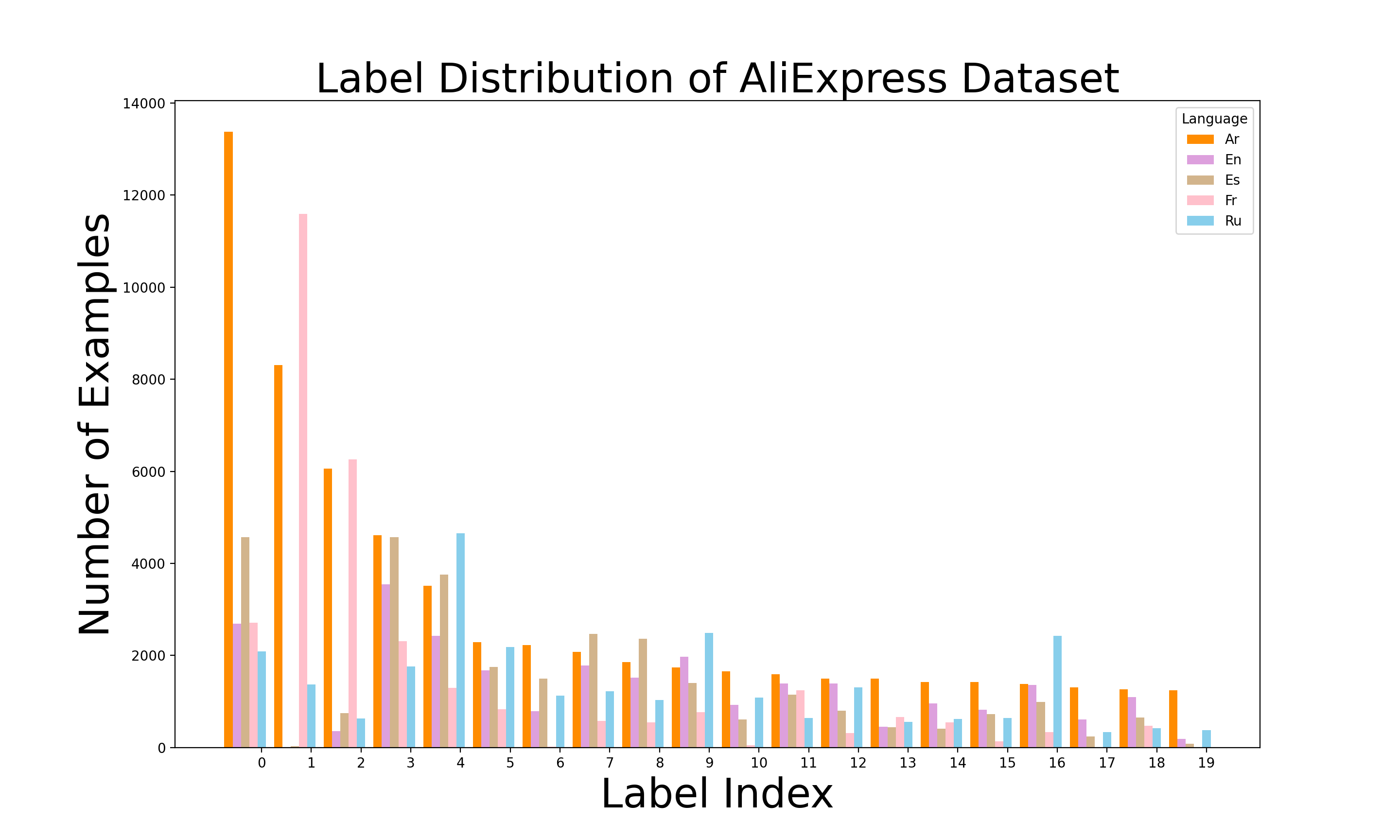}
    \caption{The label distribution of language in the practical AliExpress (AE) dataset.The number 0-19 represents the index of different categories.}
    \label{fig: langauge data distribution}
\end{figure}

\begin{figure}[!ht]
    \centering
    \includegraphics[width=1\linewidth]{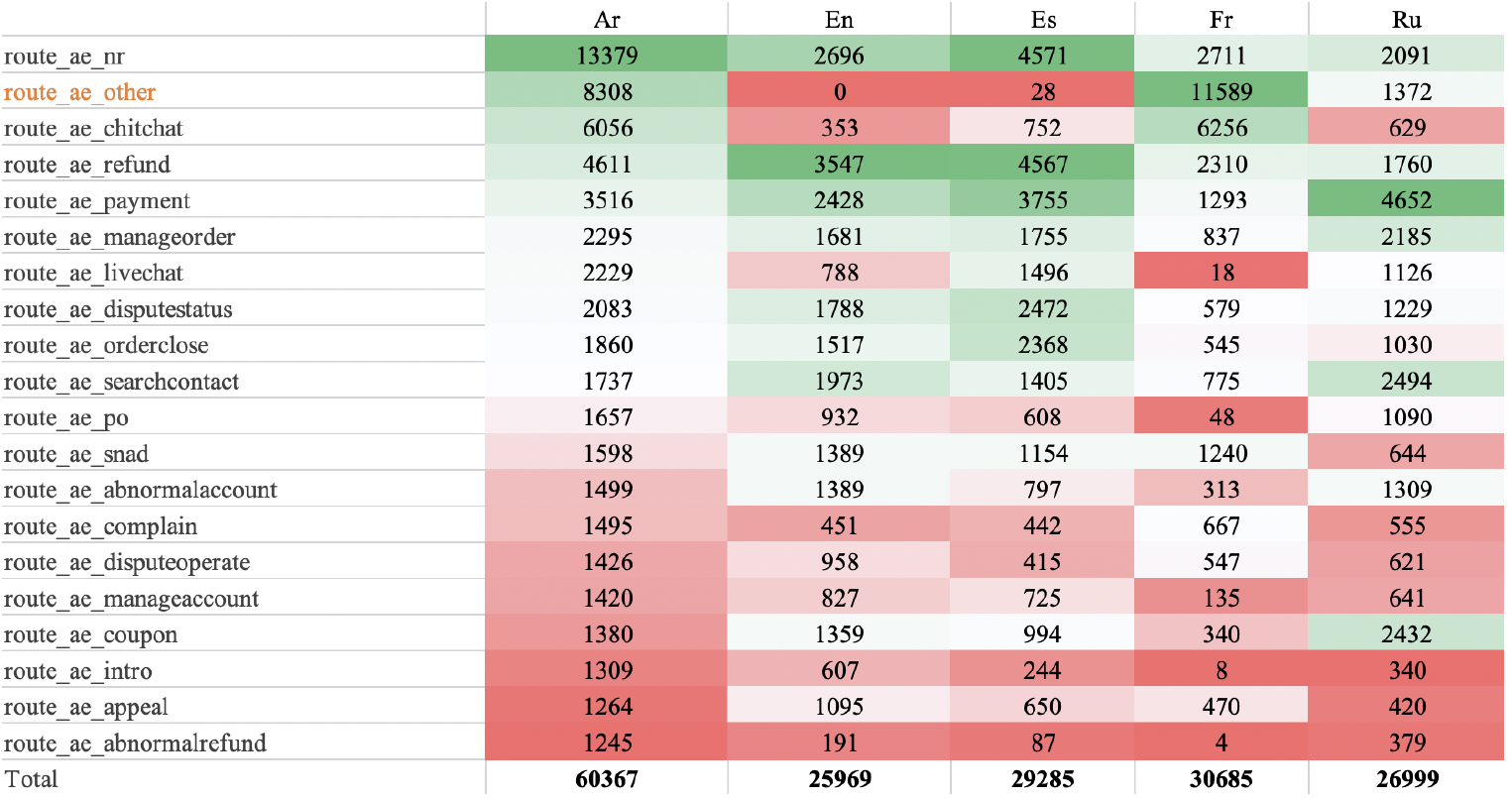}
    \caption{Statistical results of different labels in AE NLI data.}
    \label{fig:data distribution}
\end{figure}
As shown in Figure \ref{fig:data distribution}, we can observe the details of label distribution in each language. There are 20 labels in each language. However, the distribution of labels in practical are not well balanced. For example, in English set, there are 0 example in $others$. Hence, adapting to new language in practical way is a challenge for knowledge distillation.

\section{Details of Adapting to New Language }
\begin{table}[!ht]
    \centering
    \begin{tabular}{|c|c|}
    \hline
      Training Epoch  &  Ranked Index\\
    \hline
        1 & [5 6 0 2 7 4 3 1] \\
        2 & [3 0 1 4 7 2 6 5] \\
        3 & [1 6 7 0 2 3 4 5] \\
        4 & [5 4 2 3 0 6 7 1] \\
        5 & [1 7 0 3 4 6 2 5] \\
    \hline
    \end{tabular}
    \caption{The ranked index of first 5 epochs for adapting new language. Number 0-7 represent Ar, En, Es, Fr, Ru, Pt, Ko, Pl respectively. }
    \label{tab:new language}
\end{table}
The ranked results of first 5 epochs for adapting new language are shown in Table \ref{tab:new language}. We found that the top3 max-margin language number are not always 5, 6, 7 (represent Pt, Ko, Pl). As the model adapts to the new language, our strategy not only pays attention to the new language, but also maintains the accuracy of the model in the existing language, thereby reducing the degree of catastrophic forgetting of the model.


\end{document}